\documentclass[runningheads]{llncs}

\usepackage{graphicx}
\usepackage{amsmath,amssymb} 
\usepackage{color}
\usepackage{multirow}
\usepackage[width=122mm,left=12mm,paperwidth=146mm,height=193mm,top=12mm,paperheight=217mm]{geometry}

\usepackage[bookmarks=false,colorlinks=true,linkcolor=black,citecolor=black,filecolor=black]{hyperref}
\usepackage{url}
\usepackage[british,UKenglish,USenglish,english,american]{babel}

\def\eg{\emph{e.g.}}

\def\ie{\emph{i.e.}}

\newcommand{\subsubsubsection}[1]{\vspace{0.2em}\noindent\textbf{\textcolor[rgb]{0,.1,.4}{#1:}}}

\begin{document}

\title{ContextNet: Exploring Context and Detail \\for Semantic Segmentation in Real-time}

\titlerunning{~}

\authorrunning{~}

\author{Rudra P K Poudel \and Ujwal Bonde \and Stephan Liwicki \and Christopher Zach}
\institute{Toshiba Research, Cambridge, UK\\
    \email{\{rudra.poudel,ujwal.bonde,christopher.m.zach\}@gmail.com}}

\maketitle

\begin{abstract}
Modern deep learning architectures produce highly accurate results on many challenging semantic segmentation datasets. State-of-the-art methods are, however, not directly transferable to real-time applications or embedded devices, since na\"ive adaptation of such systems to reduce computational cost (speed, memory and energy) causes a significant drop in accuracy. We propose ContextNet, a new deep neural network architecture which builds on factorized convolution, network compression and pyramid representation to produce competitive semantic segmentation in real-time with low memory requirement. ContextNet combines a deep network branch at low resolution that captures global context information efficiently with a shallow branch that focuses on high-resolution segmentation details. We analyse our network in a thorough ablation study and present results on the Cityscapes dataset, achieving 66.1\% accuracy at 18.3 frames per second at full ($1024\times2048$) resolution (41.9 fps with pipelined computations for streamed data).
\end{abstract}

\section{Introduction}
\label{sec:introduction}
Semantic segmentation provides detailed pixel-level classification of images, which is particularly suited for autonomous vehicles and driver assistance, as these applications often require accurate road boundaries and obstacle detection ~\cite{cityscaples2016,kitti2015,camvid-dataset2009}. Modern systems produce highly accurate segmentation results, but often at the cost of reduced computational efficiency. In this paper, we propose ContextNet to address competitive semantic segmentation for autonomous driving tasks, requiring real-time processing and memory efficiency.

Deep neural networks (DNNs) are becoming the preferred approach for semantic image segmentation in recent years. High performance segmentation methods adopt state-of-the-art classification architecture using fully convolutional network (FCN) \cite{long2016} or encoder-decoder techniques \cite{badrinarayanan2017}. In particular, DeepLab~\cite{chen2016} employs an increased number of layers to extract complex and abstract features, leading to increased accuracy. PSPNet~\cite{zhao2017a} combines multiple levels of information through context aggregation from multiple feature resolutions, and benchmarks as one of the most accurate DNNs. The accuracy of these architectures comes at a high computational cost. Semantic segmentation of a single image requires more than a second, even on modern high-end GPUs (\eg Nvidia Titan X) and hinders their deployment for driverless cars.

Recent interest in embedded devices, wearable devices and autonomous vehicles has sparked growing focus on semantic segmentation in real-time with low energy and memory requirements \cite{badrinarayanan2017,paszke2016,zhao2017b,romera2018,hubara2016,rastegari2016,wu2018}. Common techniques include convolutional factorization \cite{howard2017,sandler2018,chollet2016}, network pruning \cite{han2016,li2017} and weight quantization \cite{hubara2016,rastegari2016,wu2018}.

\subsubsubsection{Convolution Factorization} Ordinary convolutions perform cross-channel and spatial correlations simultaneously. In contrast, convolution factorization employs multiple sub-operations to reduce computation cost and memory. Examples include Inception \cite{szegedy2016}, Xception \cite{chollet2016} and MobileNet \cite{howard2017,sandler2018}. In particular, MobileNet \cite{howard2017} decomposes a standard convolution into a depth-wise convolution (also known as spatial or channel-wise convolution) and a $1\times1$ point-wise convolution. MobileNetV2 \cite{sandler2018} further improves this framework by introducing a bottleneck block and residual connections \cite{he2015}.

\subsubsubsection{Network Compression} Network compression is orthogonal to convolution factorization. Network hashing or pruning is applied to reduce the size of a pre-trained network, resulting in faster test-time execution and a smaller parameter set and memory footprint~\cite{han2016,li2017}.

\subsubsubsection{Network Quantization} The runtime of a network can be further reduced using quantization techniques \cite{hubara2016,rastegari2016,wu2018}. These techniques reduce the size/memory requirements of a network by encoding parameters in low-bit representations. Moreover, runtime is further improved if binary weights and activation functions are employed since efficient XNOR and bit-count operations replace costly floating point operations of standard DNNs.

Despite these known techniques, notably, only ENet \cite{paszke2016} implements a network that runs in real-time on the full resolution images of the popular Cityscapes dataset \cite{cityscaples2016}, but with significantly reduced accuracy in comparison to the state-of-the-art.

\subsection{Contributions}
\begin{figure}[t]
\centering
\includegraphics[width=1\linewidth]{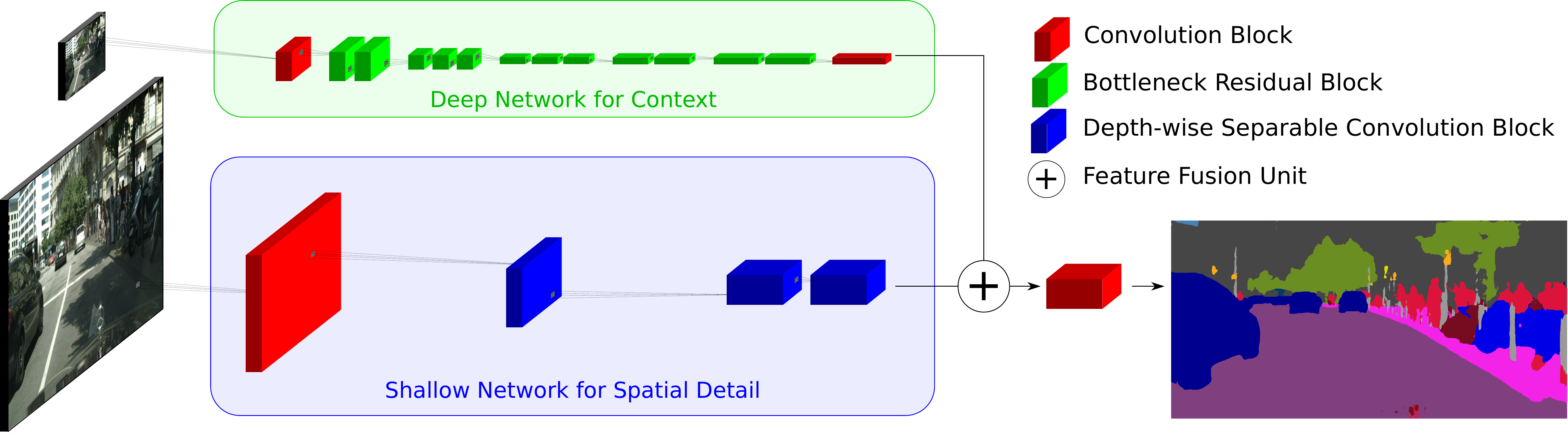}
\caption{ContextNet combines a deep network at small resolution with a shallow network at full resolution to achieve accurate and real-time semantic segmentation.}\label{fig:teaser}
\end{figure}
In this paper we introduce ContextNet, a competitive network for semantic segmentation running in real-time with low memory footprint (Figure~\ref{fig:teaser}). ContextNet builds on the following two observations of previous work:
\begin{enumerate}
  \item{An increased number of layers helps to learn more complex and abstract features, leading to increased accuracy but also increased running time \cite{he2015,chen2016,zhao2017a}.}
  \item{Aggregation of context information from multiple resolutions is beneficial, since it combines multiple levels of information for improved performance \cite{long2016,zhao2017a}.}
\end{enumerate}
Consequently, we propose a network architecture with branches at two resolutions. In order to obtain real-time performance, the network at low resolution is deep, while the network at high resolution is rather shallow. As such, our network perceives context at low resolution and refines it for high resolution results. The architecture is designed to ensure the low resolution branch is mainly responsible for providing the correct label, whereas the high-resolution branch refines the segmentation boundaries.

Our design is related to pyramid representations \cite{burt1987} which have been employed recently in many DNNs. RefineNet \cite{lin2017} uses multiple paths over which information from different resolutions is carefully combined to obtain high-resolution segmentation results. Ghiasi and Fowlkes \cite{ghiasi2016} use a class specific multi-level Laplacian pyramid reconstruction technique for the segmentation task. However, neither achieve real-time performance.

ICNet \cite{zhao2017b} employs a subset of ResNet \cite{he2015} at three resolution levels (full, half and one fourth of the original input resolution) which are later combined to provide semantic segmentation results. Here we emphasize, while our implementation of ICNet confirms accuracy in \cite{zhao2017b}, the reported runtime is achieved only at half resolution images of Cityscapes~\cite{cityscaples2016}. In contrast, we experimentally show that it is possible to capture global context (semantically rich features) with only a single deeper network on smaller input size and local context (detailed spatial features) with a shallow network on full resolution.

We employ efficient \textit{bottleneck residual blocks}~\cite{sandler2018} and \textit{network pruning}~\cite{han2016,li2017} to present a novel DNN architecture for real-time semantic segmentation with full floating point operations. Network quantization is not explored and is left for future work. In the following sections, we provide design choices of our proposed ContextNet and describe our detailed ablation analysis and experiments on Cityscapes~\cite{cityscaples2016}.

\section{Proposed Context Network (ContextNet)}
\label{sec:context-net}
The proposed ContextNet is visualized in Figure~\ref{fig:teaser}. ContextNet produces cost efficient accurate segmentation at low resolution, which is then combined with a sub-network at high resolution to provide detailed segmentation results.

\subsection{Motivation}
Combining different levels of context information is advantageous for the semantic segmentation task~\cite{ronneberger2015,long2016,chen2016}. PSPNet \cite{zhao2017a} employs an explicit pyramid pooling module to improve performance by capturing global and local context at different feature resolutions.

Another noticeable trend is that state-of-the-art DNNs have grown deeper (\eg \cite{he2015,zhao2017a,chen2016}), since this can capture more complex and abstract features, and increase the receptive field. Unfortunately, higher number of layers ultimately increase runtime and memory requirements.

ContextNet combines both, deep networks and multi-resolution architectures. In order to achieve fast runtime we restrict our multi-scale input to two branches, where global information at low resolution is refined by a shallow network at high resolution to produce the final segmentation results in real-time.

\subsection{Network Design}\label{ssec:net-architecture}
We now describe the main building blocks, the overall architecture and discuss our design.

\subsubsection{Depth-wise Convolution to Improve Run-time}
\begin{table}[t]
\begin{center}
\scalebox{0.8}{%
        \begin{tabular}{|c|c|c|} \hline
        Input & Operator& Output \\ \hline
        $h \times w \times c$  & Conv2d $1/1,f$  & $h \times w \times tc$ \\
        $h \times w \times tc$ & DWConv $3/s,f$ & $\frac{h}{s} \times \frac{w}{s} \times tc$ \\
        $\frac{h}{s} \times \frac{w}{s} \times tc$ & Conv2d $1/1,-$ & $\frac{h}{s} \times \frac{w}{s} \times {c}'$ \\
        \hline
\end{tabular}}
\end{center}
\caption{\textit{Bottleneck residual block} transferring the input from $c$ to ${c}'$ channels with $h$ height, $w$ width, expansion factor $t$, convolution type kernel-size/stride $s$ and non-linear function $f$.}\label{tbl:bottleneck-residual-block}
\end{table}
Depth-wise separable convolutions factorize standard convolution (Conv2d) into a depth-wise convolution (DWConv), also known as spatial or channel-wise convolution, followed by a $1\times1$ point-wise convolution layer \cite{howard2017}. Cross-channel and spatial correlation is therefore computed independently, which drastically reduces the number of parameters, resulting in fewer floating point operations and fast execution time.

ContextNet utilizes DWConv, as we design its two main building blocks accordingly (Figure \ref{fig:teaser}). The sub-network with down-sampled input uses bottleneck residual blocks with DWConv~\cite{sandler2018} (Table~\ref{tbl:bottleneck-residual-block}). In the sub-network at full resolution depth-wise separable convolutions are directly employed. We omit the nonlinearity between depth-wise and point-wise convolutions in our full resolution branch, since it had limited impact on accuracy in our initial experiments.

\subsubsection{Capturing Global and Local Context}
ContextNet has two branches, one for full resolution ($h \times w$) and one for lower resolution (\eg $\frac{h}{4} \times \frac{w}{4}$), with input image height $h$ and width $w$ (Figure~\ref{fig:teaser}). Each branch has different responsibilities; the latter captures the global context of the image, and the former provides the detail information for the higher resolution segmentation. In particular, our design choices are motivated as follows:
\begin{enumerate}
  \item For fast feature extraction, semantically rich features are extracted only from the lowest possible resolution.
  \item Features for local context are extracted separately from full resolution input by a very shallow branch, and are then combined with low-resolution results.
\end{enumerate}
Hence, significantly faster computation of image segmentation is possible in ContextNet.
\begin{table}[t]
\begin{center}
\scalebox{0.8}{%
       \begin{tabular}{|c|c|c|c|c|c|} \hline
        Input & Operator & Expansion Factor & Output Channels & Repeats & Stride \\ \hline
        $256\times512\times3$  & Conv2d     & -  & 32  & 1 & 2 \\ \hline
        $128\times256\times32$ & bottleneck & 1  & 32  & 1 & 1 \\ \hline
        $128\times256\times32$ & bottleneck & 6  & 32  & 1 & 1 \\ \hline
        $128\times256\times32$ & bottleneck & 6  & 48  & 3 & 2 \\ \hline
        $64\times128\times48$   & bottleneck & 6  & 64  & 3 & 2 \\ \hline
        $32\times64\times64$   & bottleneck & 6  & 96  & 2 & 1 \\ \hline
        $32\times64\times96$   & bottleneck & 6  & 128 & 2 & 1 \\ \hline
        $32\times64\times128$   & Conv2d & -  & 128 & 1 & 1 \\ \hline
        \end{tabular}}
\end{center}
\caption{Branch-4 for compressed input. Repeated block use stride 1 after first block/layer.}\label{tbl:contextnet-branch4-model}
\end{table}

\subsubsubsection{Capturing context} The detail structure of the lower resolution branch is shown in Table~\ref{tbl:contextnet-branch4-model}. This sub-network consists of two convolution layers and 12 bottleneck residual blocks. Similar to MobileNetV2 \cite{sandler2018}, we employ residual connections for bottleneck residual blocks when input and output are of the same size. While the low resolution branches of ICNet \cite{zhao2017b} requires 50 costly layers of ResNet \cite{he2015}, in ContextNet a total of only 38 highly efficient layers are used to describe global context.

\subsubsubsection{Spatial detail} The sub-network of the full resolution branch is kept as shallow as possible and only consists of four layers. Its objective is to refine the results with local context. In particular, the number of feature maps are 32, 64, 128 and 128 respectively. The first layer uses standard convolution while all other layers use depth-wise separable convolutions with kernel size $3\times3$. The stride is 2 for all but the last layer, where it is 1.

\begin{table}[t]
\begin{center}
\scalebox{0.8}{%
        \begin{tabular}{|c|c|} \hline
        Branch-1 & Branch-4 \\ \hline
        \hspace{5.5em}-\hspace{5.5em}~& Upsample $\times$ 4 \\
        - & DWConv (dilation 4) $3/1,f$ \\
        Conv2d $1/1,-$ &  Conv2d $1/1,-$ \\ \hline
        \multicolumn{2}{|c|}{add,$f$} \\
        \hline
\end{tabular}}
\end{center}
\caption{Features fusion unit of ContextNet.}
\label{tbl:feature-fusion-unit}
\end{table}
We use fusion unit shown in Table~\ref{tbl:feature-fusion-unit} to merge the features from both branches. Since runtime is of concern, we use feature addition instead of concatenation. Finally, we use a simple $1 \times 1$ convolution layer for the final soft-max based classification results.

\subsubsection{Design Choices}
We have conducted several initial experiments before deciding on the final model of ContextNet using train and validation sets of the Cityscapes dataset. We have found that the use of a pyramid pooling module~\cite{zhao2017a} after the low-resolution branch increases accuracy. Also, learning global context using down-sampled input is more efficient than learning with asymmetric convolution (for examples, $k \times 1$, $1 \times k$, where $k=5/7/9$) on higher resolution inputs. Class weight balancing technique did not help, when we increased batch-size to 16 or more.

Empirically, we found a weighted auxiliary loss for the low-resolution branch is beneficial. We think, the auxiliary loss ensures that meaningful features for semantic segmentation are extracted by the branch for global context, and are learned independently from the other branch. The weight of the auxiliary loss was set to 0.4. Following \cite{chen2016,zhao2017a}, a cross-entropy loss is employed as auxiliary and final loss of ContextNet.

\section{Experiments}\label{sec:experiments}
In our evaluation, we present a detailed ablation study of ContextNet on the validation set of the Cityscapes dataset \cite{cityscaples2016} , and report its performance on the Cityscapes benchmark.

\subsection{Implementation Details}\label{ssec:implementation-details}
All our experiments are performed on a workstation with Nvidia Titan X (Maxwell, 3072 CUDA cores), CUDA 8.0 and cuDNN V6. We use ReLU6 as nonlinearity function due to its robustness when used with low-precision computations \cite{howard2017}. During training, batch normalization is used at all layers and dropout is used before the soft-max layer only. During inference, parameters of batch normalization are merged with the weights and bias of parent layers. In the depth-wise convolution layers, we found that $\ell_2$ regularization is not necessary, which is consistent with the findings in \cite{howard2017}. Since labelled training data is limited, we apply standard data augmentation techniques in all experiments: random scale 0.5 to 2, horizontal flip, varied hue, saturation, brightness and contrast.

The models of ContextNet are trained with TensorFlow \cite{tensorflow2015} using RMSprop \cite{tieleman2012} with a discounting factor of 0.9, momentum of 0.9 and epsilon parameter equal to~1. Additionally, we apply a \textit{poly learning rate}~\cite{chen2016} with base rate 0.045 and power 0.98. The maximum number of epochs is set to 1000, as no pre-training is used.

Results are reported as mean intersection-over-union (mIoU) \cite{cityscaples2016} and runtime considers single threaded CPU with sequential CPU to GPU memory transfer, kernel execution, and GPU to CPU memory exchange.

\subsection{Cityscapes Dataset}
Cityscapes is a large-scale dataset for semantic segmentation that contains a diverse set of images in street scenes from 50 different cities in Germany \cite{cityscaples2016}. In total, it consists of 25,000 annotated $1024\times2048 px$ images of which 5,000 have labels at high pixel accuracy and 20,000 are weakly annotated. In our experiments we only use the 5,000 images with high label quality: a training set of 2,975 images, validation set of 500 images and 1,525 test images which can be evaluated on the Cityscapes evaluation server. No pre-training is employed.

\subsubsection{Ablation Study}\label{ssec:ablation-study}
In our ablation study weights are learned solely from the Cityscapes training set, and we report the performance on the validation set. In particular, we present effects on different resolution factors of the low resolution branch, introducing multiple branches, and modifications on the number of parameters. Finally, we analyse the two branches in detail.

\begin{table}[t]
\begin{center}
\scalebox{0.8}{%
  \begin{tabular}{|l|c|c|c||c||c|c|} \hline
  ~ & \textbf{cn18} & \textbf{cn14} & \textbf{cn12} & \textbf{cn124} & \textbf{cn14-500} & \textbf{cn14-160}\\ \hline
        Accuracy (mIoU in \%) & 60.1 & 65.9 & 68.7 & 67.3 & 62.1 & 57.7 \\ \hline
        Frames per Second & 21.0 & 18.3 & 11.4 & 7.6 & 20.6 & 22.0 \\ \hline
        Number of Parameters (in millions) & 0.85 & 0.85 & 0.85 & 0.93 & 0.49 & 0.16\\ \hline
        \end{tabular}}
    \end{center}
    \caption{ContextNet (cn14) compared to its version with half resolution (cn12) and eighth resolution (cn18) at the low resolution branch, and with multiple levels at quarter, half and full resolution (cn124) on Cityscapes validation set. Implementations with smaller memory footprint are also shown (cn14-500 and cn14-160).} \label{tbl:ablation-num-branches}
\end{table}

\subsubsubsection{Input resolution} The input image resolution is the most critical factor for the computation time. Our low resolution branch takes images of a quarter size at $256\times512 px$ for Cityscapes images (denoted cn14). Alternatively, half (cn12) or one eighth (cn18) resolution could be used. Table~\ref{tbl:ablation-num-branches} shows how the different options affect the results. Overall, larger resolution in the deeper context branch produce better segmentation results. However, these improvements come at the cost of running time.

Table~\ref{tbl:ablation-sub-input-size-results} lists the IoU in more detail. As expected, accuracy of small-size classes (\ie fence, pole, traffic light and traffic sign), classes with fine detail (\ie person, rider, motorcycle and bicycle) and rare classes (\ie bus, train and truck) benefit from increased resolution at the global context branch. Other classes are often of larger size, and can therefore be captured at lower resolution. We conclude that cn14 is fairly balanced at 18.3 frames per seconds (fps) and 65.9\% mIoU.
\begin{table}[t]
\begin{center}
\resizebox{\textwidth}{!}{%
    \begin{tabular}{|l|c|c|c|c|c|c|c|c|c|c|c|c|c|c|c|c|c|c|c|} \hline
    & \rotatebox[origin=c]{90}{road} & \rotatebox[origin=c]{90}{sidewalk} & \rotatebox[origin=c]{90}{building} & \rotatebox[origin=c]{90}{wall} & \rotatebox[origin=c]{90}{\textcolor[rgb]{0, 0.7, 0}{\textbf{fence}}} & \rotatebox[origin=c]{90}{\textcolor[rgb]{0, 0.7, 0}{\textbf{pole}}} & \rotatebox[origin=c]{90}{\textcolor[rgb]{0, 0.7, 0}{\textbf{traffic light}}} & \rotatebox[origin=c]{90}{\textcolor[rgb]{0, 0.7, 0}{\textbf{traffic sign}}} & \rotatebox[origin=c]{90}{vegetation} & \rotatebox[origin=c]{90}{terrain} & \rotatebox[origin=c]{90}{sky} & \rotatebox[origin=c]{90}{\textcolor[rgb]{0, 0, 1}{\textbf{person}}} & \rotatebox[origin=c]{90}{\textcolor[rgb]{0, 0, 1}{\textbf{rider}}} & \rotatebox[origin=c]{90}{car} & \rotatebox[origin=c]{90}{\textcolor[rgb]{1, 0, 0}{\textbf{truck}}} & \rotatebox[origin=c]{90}{\textcolor[rgb]{1, 0, 0}{\textbf{bus}}} & \rotatebox[origin=c]{90}{\textcolor[rgb]{1, 0, 0}{\textbf{train}}} & \rotatebox[origin=c]{90}{\textcolor[rgb]{0, 0, 1}{\textbf{~~motorcycle~~}}} & \rotatebox[origin=c]{90}{\textcolor[rgb]{0, 0, 1}{\textbf{bicycle}}} \\ \hline
    cn18 & 96.2 & 72.4 & 87.4 & 45.9 & 44.1 & 32.1 & 38.4 & 54.3 & 87.9 & 54.1 & 91.0 & 62.3 & 36.1 & 88.3 & 57.6 & 59.1 & 34.2 & 41.2 & 58.4  \\ \hline
    cn14 & 96.8 & 76.6 & 88.6 & 46.4 & 49.7 & 38.0 & 45.3 & 60.5 & 89.0 & 59.3 & 91.4 & 67.5 & 41.7 & 90.0 & 63.5 & 71.7 & 57.1 & 41.5 & 64.6 \\ \hline
    cn12 & 97.2 & 78.9 & 89.2 & 47.2 & 54.4 & 39.5 & 55.3 & 63.8 & 89.4 & 59.8 & 91.5 & 70.2 & 47.5 & 91.1 & 70.2 & 76.2 & 63.7 & 51.36 & 67.8 \\ \hline\hline
    cn124 & 97.4 & 79.6 & 89.5 & 44.1 & 49.8 & 45.5 & 50.6 & 64.6 & 90.2 & 59.4 & 93.4 & 70.9 & 43.1 & 91.8 & 65.2 & 71.9 & 64.5 & 41.95 & 66.1 \\ \hline
    \end{tabular}}
\end{center}
      \caption{Detailed IoU of ContextNet (cn14) compared to version with half (cn12) and eighth (cn18) resolution, and its multi-level version (cn124). Small-size classes (\textcolor[rgb]{0,0.7,0}{green}), classes with fine detail (\textcolor[rgb]{0,0,1}{blue}), and classes with very few samples (\textcolor[rgb]{1,0,0}{red}) benefit from high resolution.}
  \label{tbl:ablation-sub-input-size-results}
\end{table}

\subsubsubsection{Multiple Branches} We designed ContextNet under consideration of runtime using only two branches. However, branches at multiple resolutions could be employed. In addition to previous results, Table~\ref{tbl:ablation-num-branches} and Table~\ref{tbl:ablation-sub-input-size-results} also include a version of ContextNet with two shallow branches at half and full resolution (cn124). In comparison to cn14, cn124 improves accuracy by 1.4\% which confirms earlier results on multi-scale feature fusion \cite{lin2017,ghiasi2016,zhao2017b}. Runtime however is reduced more than twice from 18.3 fps to 7.6 fps. Furthermore we note, cn12 which has a deep sub-network at half resolution outperforms cn124 in terms of accuracy and speed. These results show that using deep sub-network on higher resolution is more beneficial than on lower resolution. Further, the number of layers have positive effect on accuracy and negative effect on run-time. We therefore conclude that a two branch architecture is most promising in our implementation of ContextNet to run in real-time.

\subsubsubsection{Number of Parameters} Apart from runtime, memory footprint is an important consideration for implementations on embedded devices. Table~\ref{tbl:ablation-num-branches} includes two versions of ContextNet with drastically reduced number of parameters, denoted as cn14-500 and cn14-160. Surprisingly, ContextNet with only 159,387 and 490,459 parameters achieved 57.7\% and 62.1\% mIoU, respectively, on the Cityscapes validation set. These results show that the ContextNet design is flexible enough to adapt the computation time and memory requirements of the given system.

\subsubsubsection{ContextNet vs Ensemble Nets}
Global context with one fourth resolution and detail branches trained independently obtained 63.3\% and 25.1\% mIoU respectively. As expected, we found that context branch is not performing well on small-size classes and receptive field of the detail branch is not large enough for reasonable performance. Outputs (softmax) of context and detail networks are averaged to create an ensemble of networks, which are trained independently. Ensemble of both branches obtained 60.3\% mIoU, which is 6.6\% less than cn14 and 3\% less than using the context branch alone. This provides further evidence that ContextNet architecture is a better choice for multi-scale features fusion.

\begin{figure}[t!]
\begin{center}
    \includegraphics[width=0.32\textwidth]{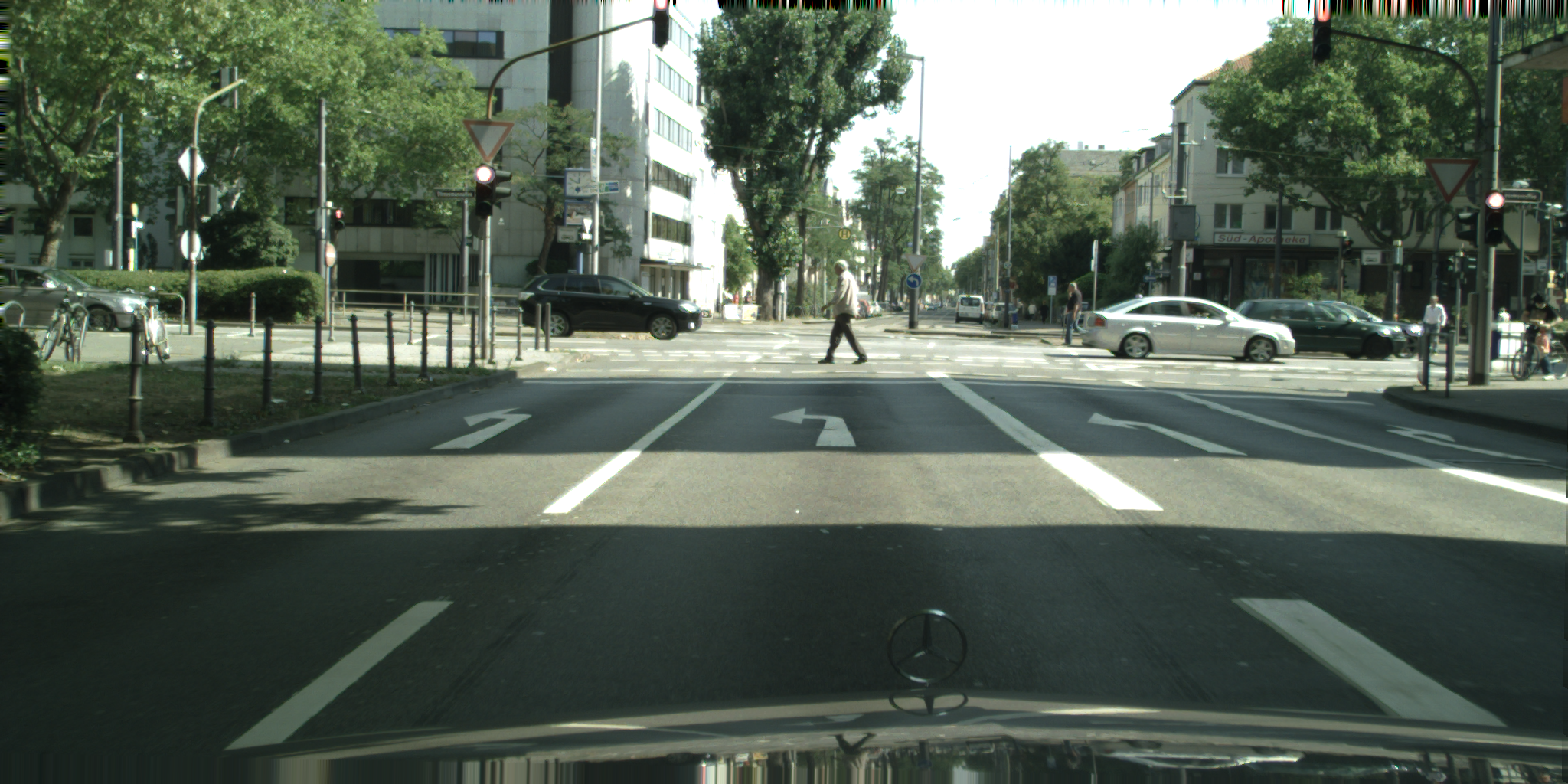}
    \includegraphics[width=0.32\textwidth]{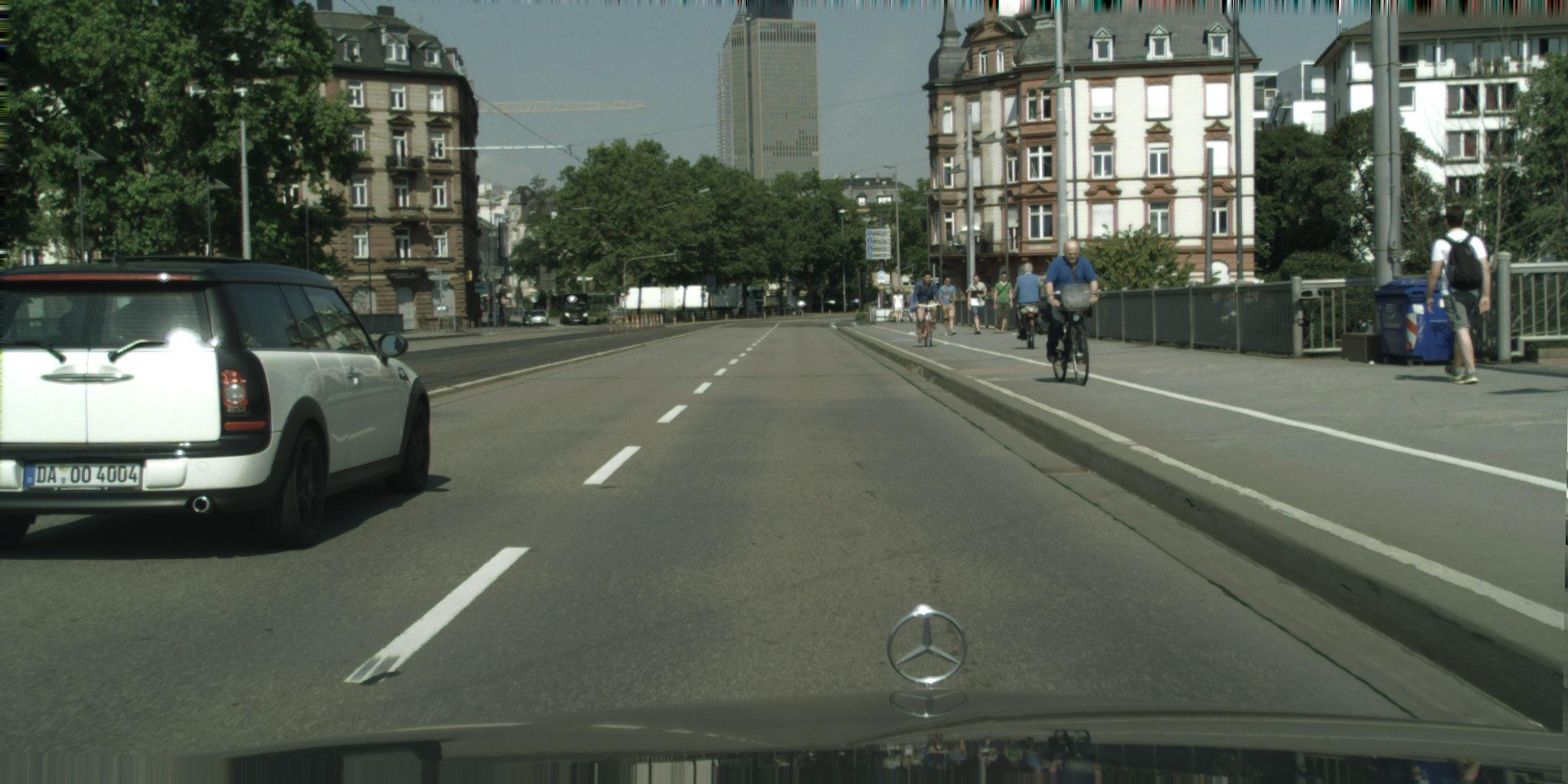}
    \includegraphics[width=0.32\textwidth]{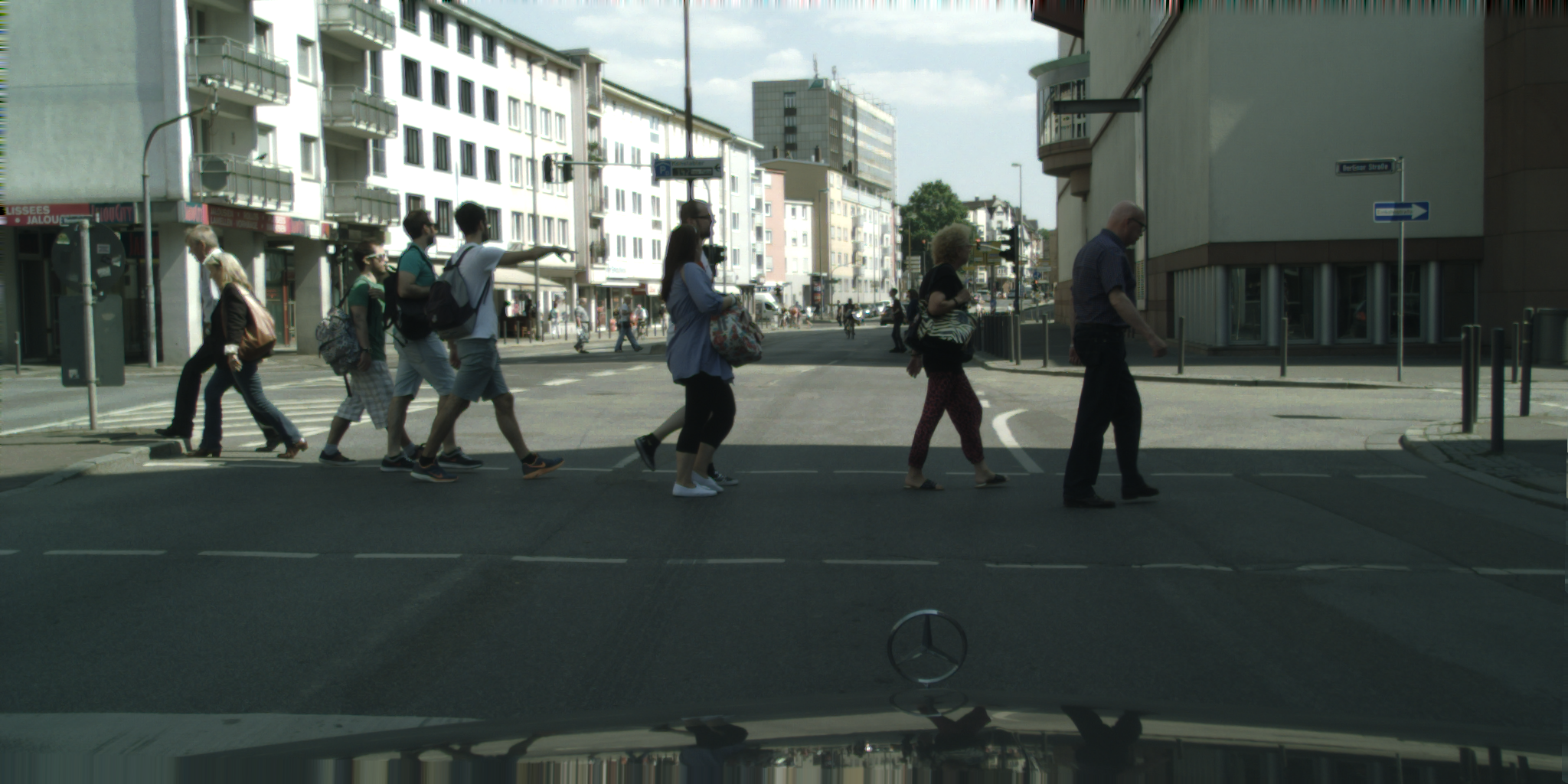}
    \\
    \includegraphics[width=0.32\textwidth]{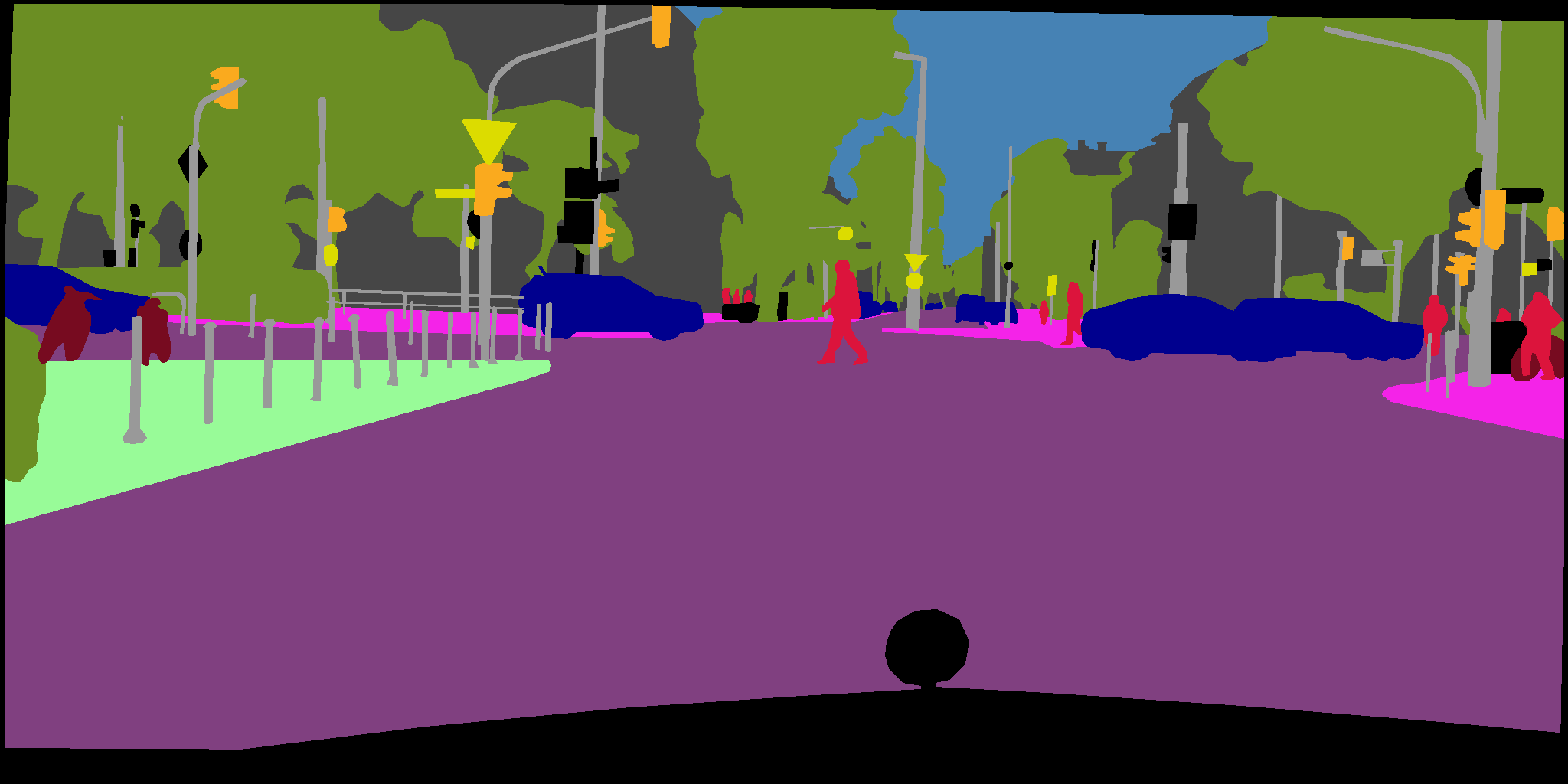}
    \includegraphics[width=0.32\textwidth]{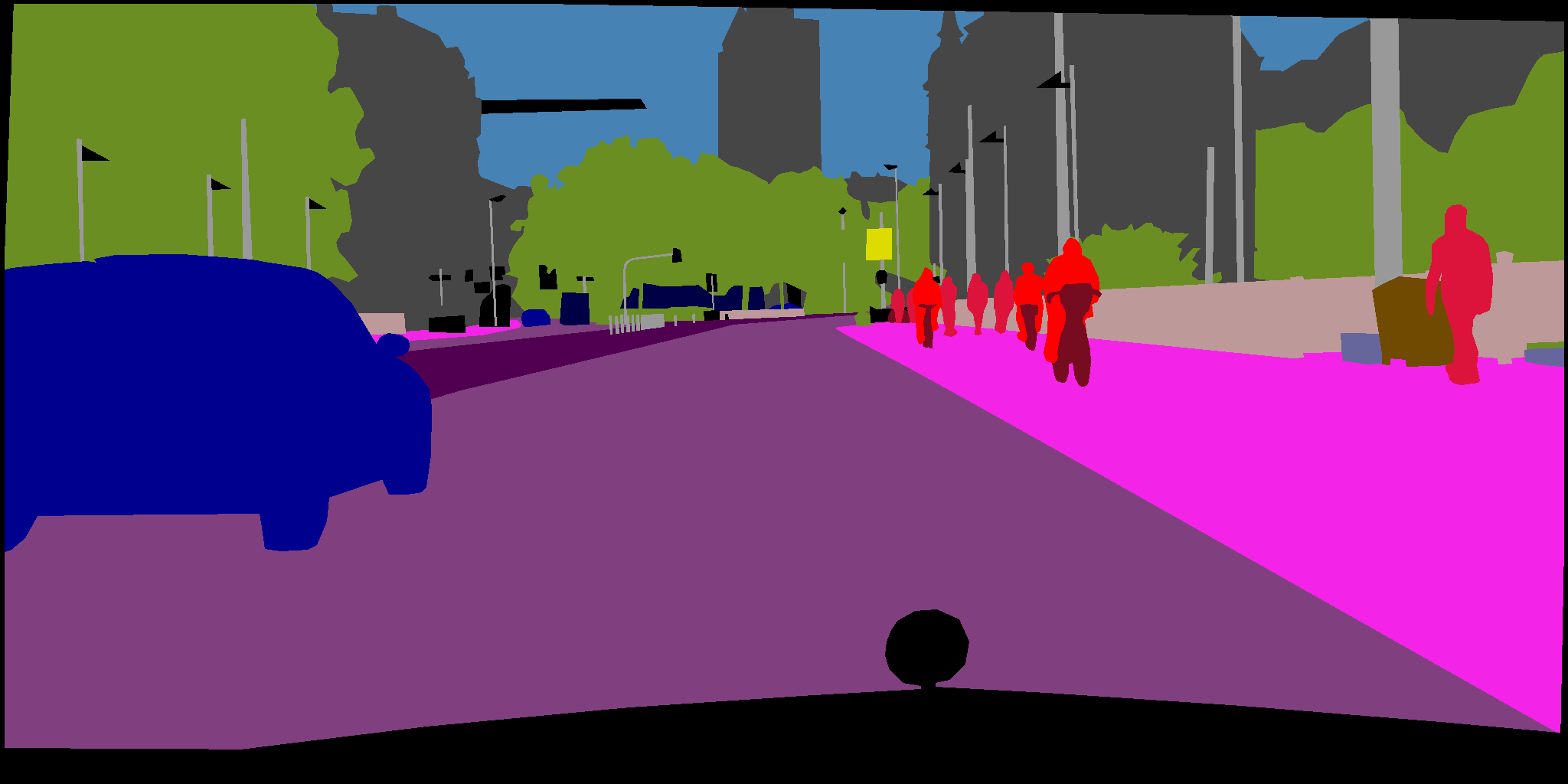}
    \includegraphics[width=0.32\textwidth]{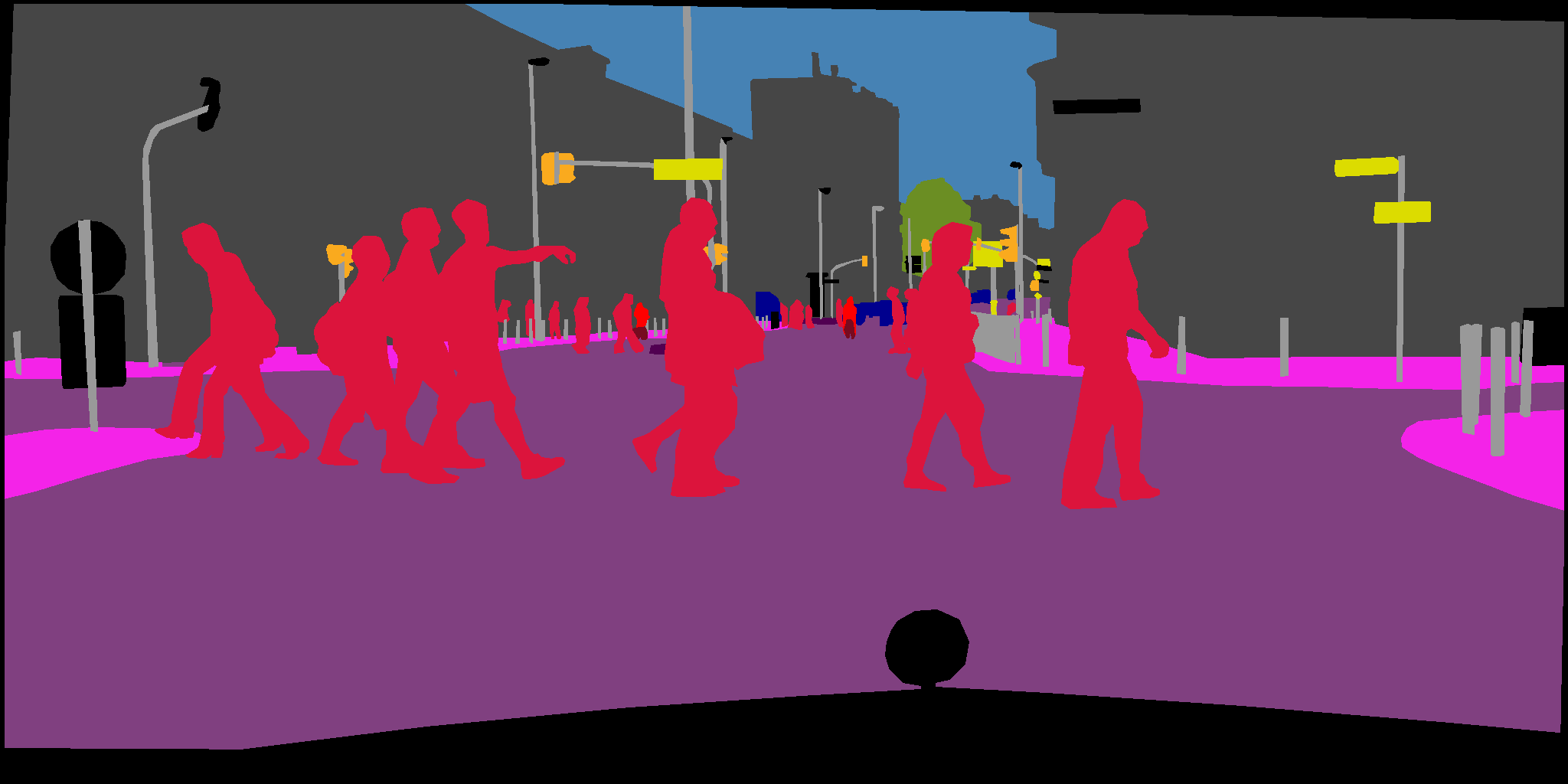}
    \\
    \includegraphics[width=0.32\textwidth]{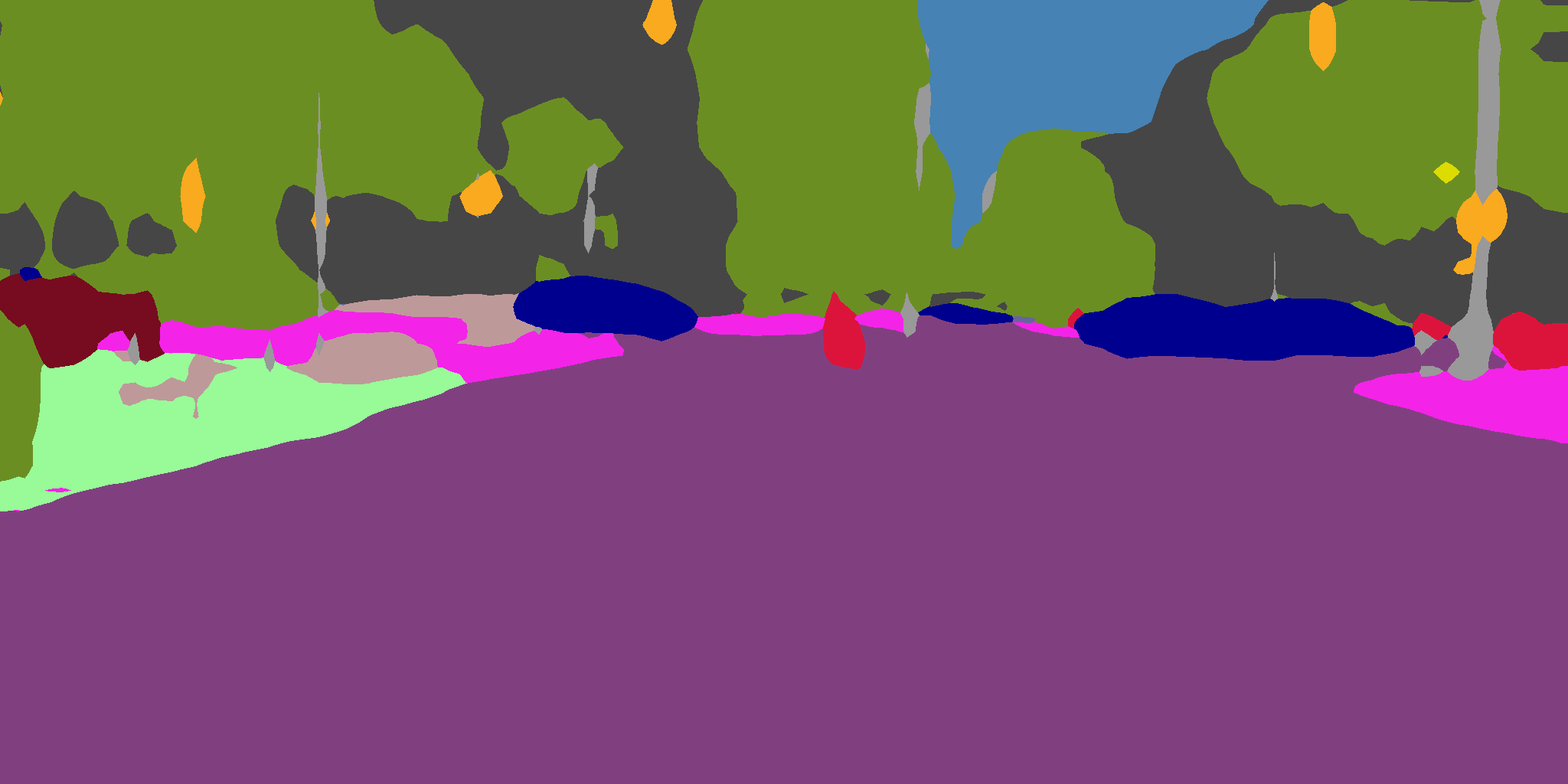}
    \includegraphics[width=0.32\textwidth]{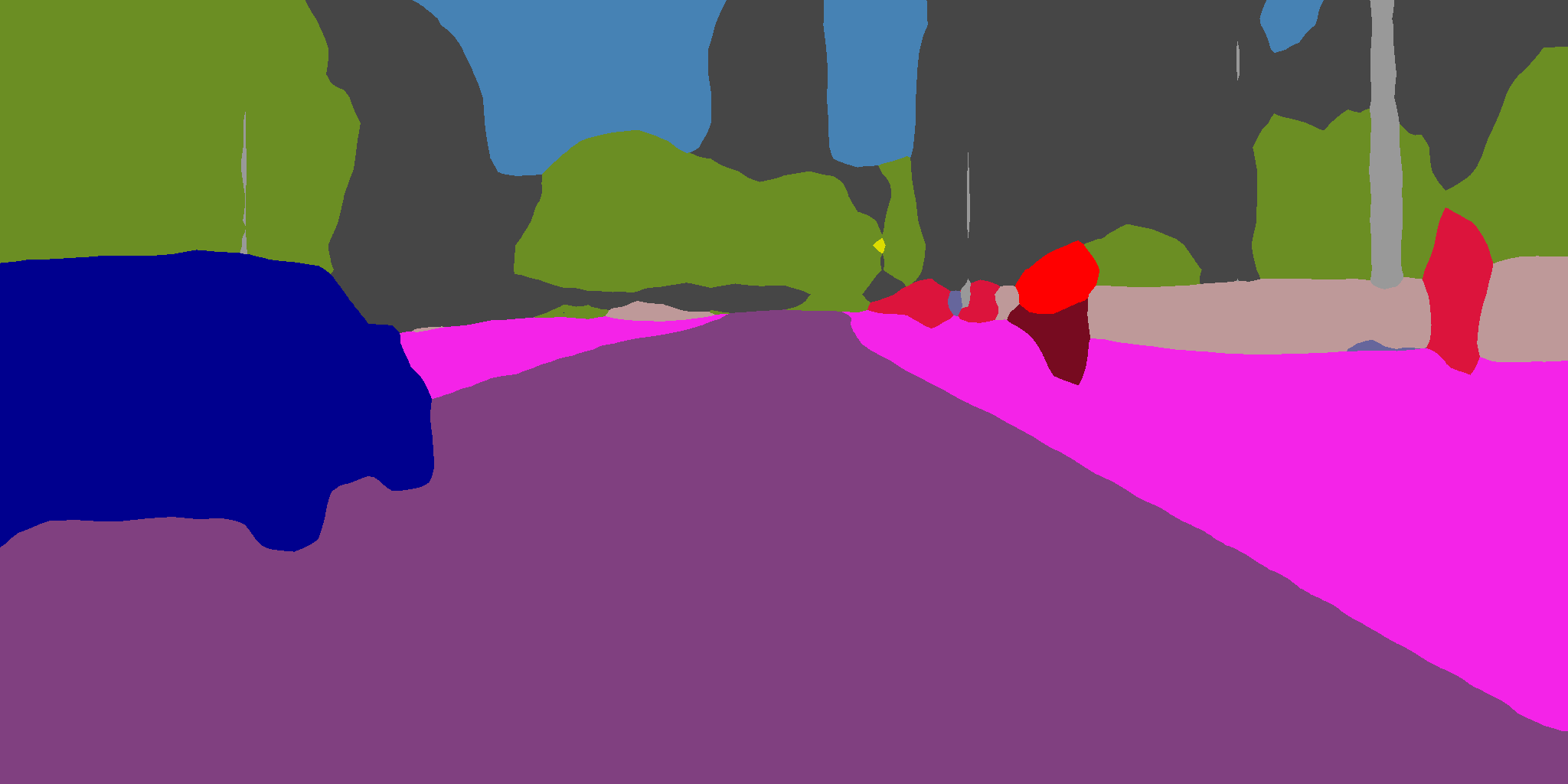}
    \includegraphics[width=0.32\textwidth]{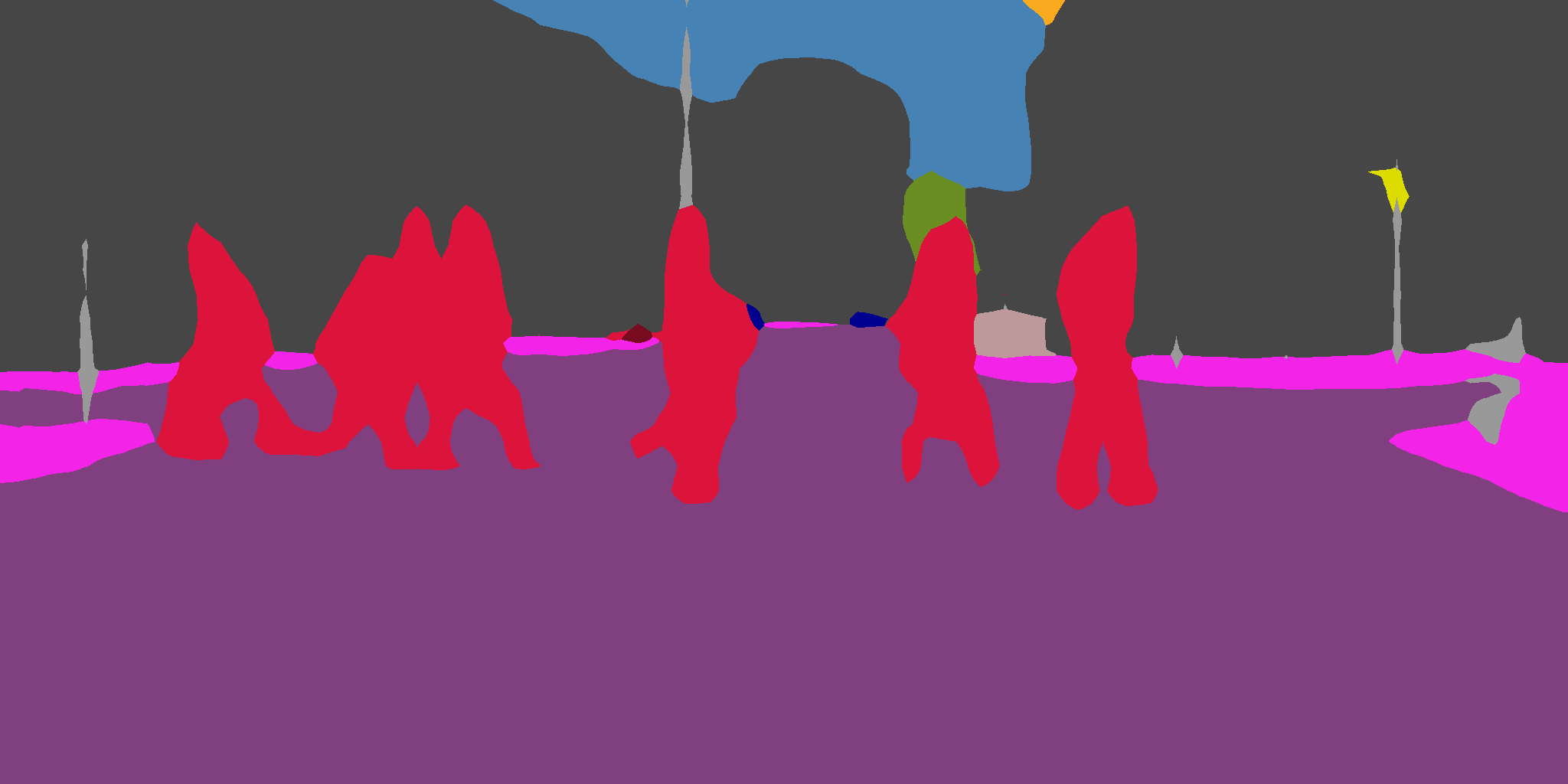}
    \\
    \includegraphics[width=0.32\textwidth]{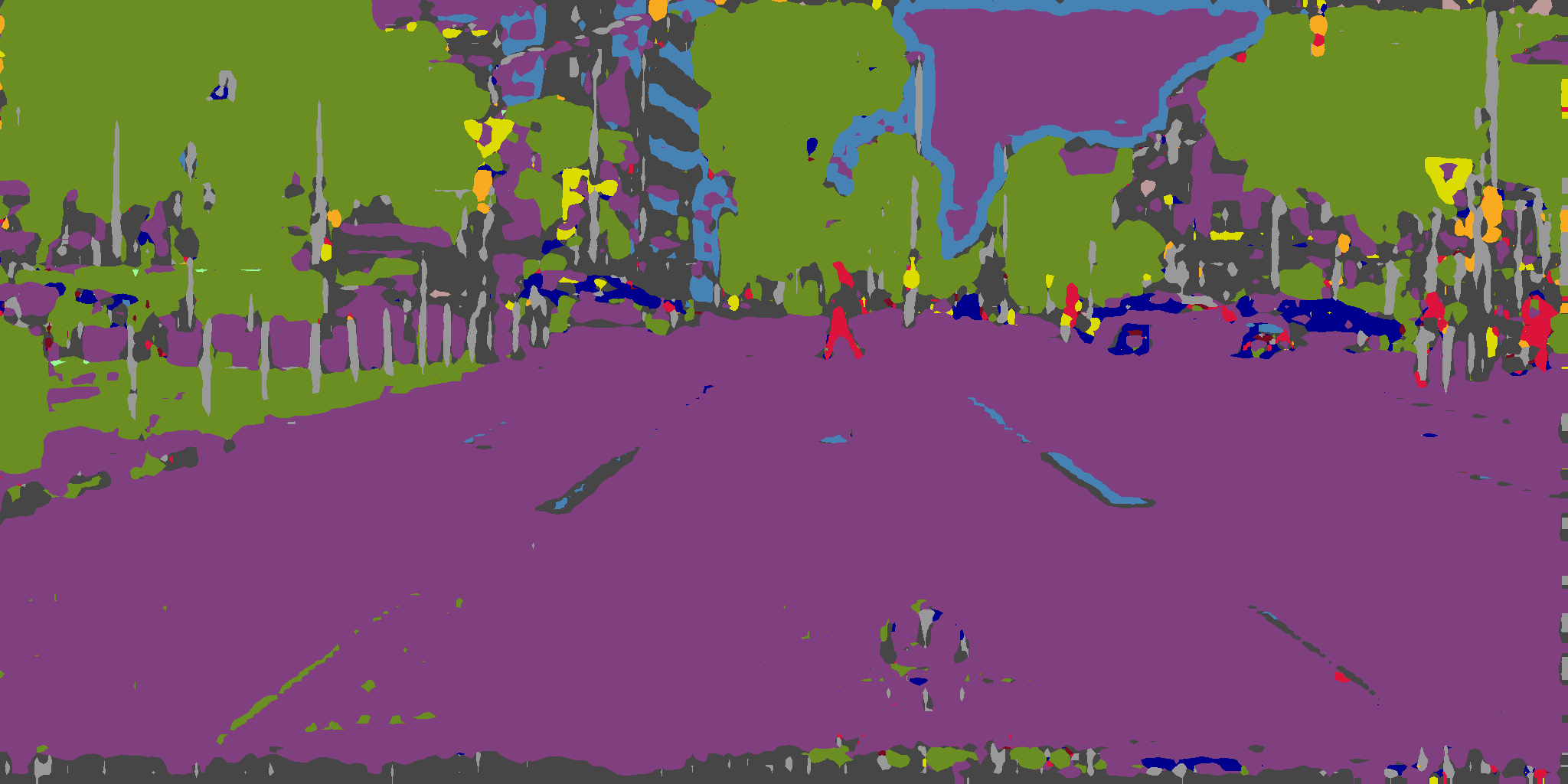}
    \includegraphics[width=0.32\textwidth]{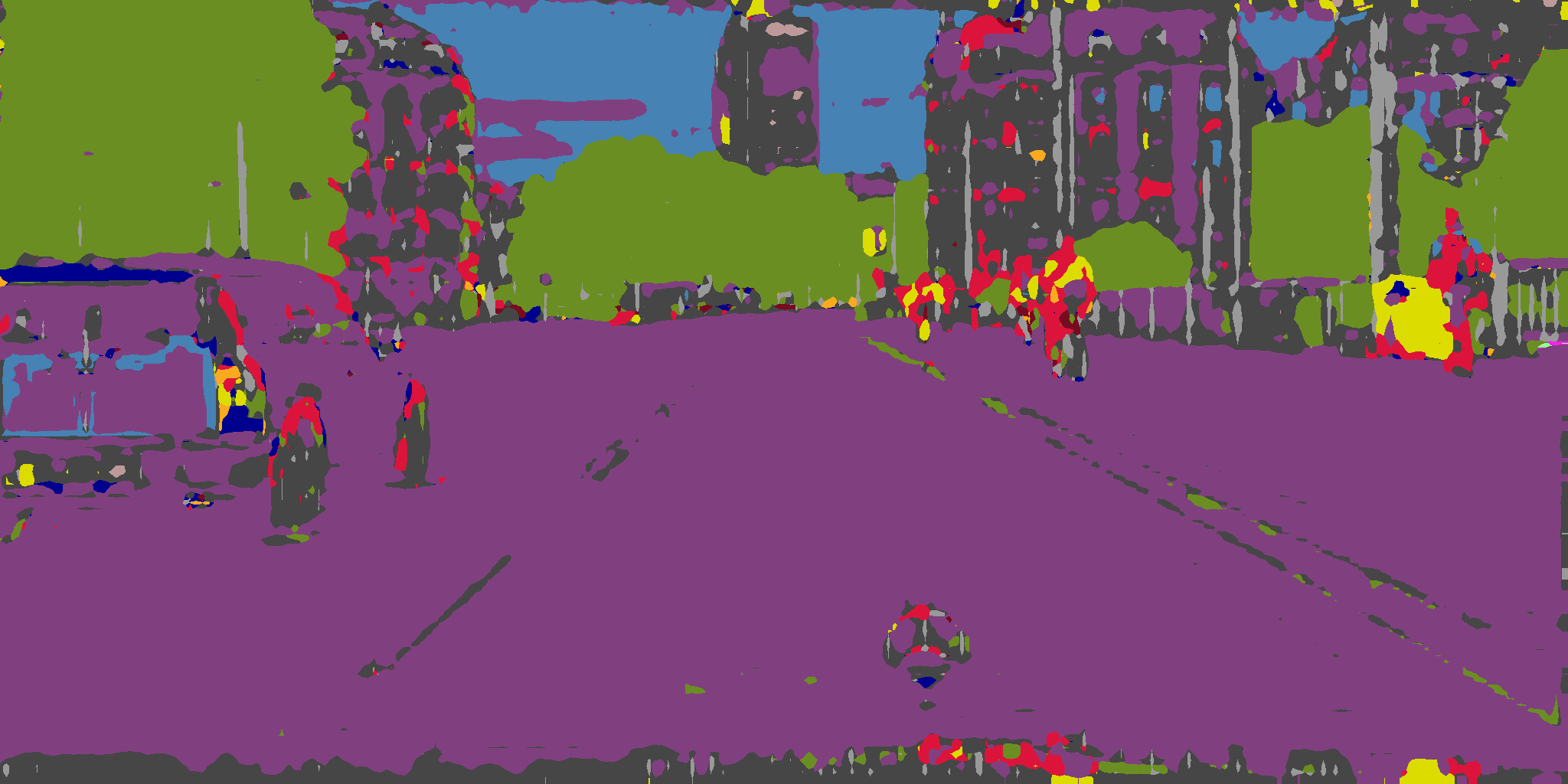}
    \includegraphics[width=0.32\textwidth]{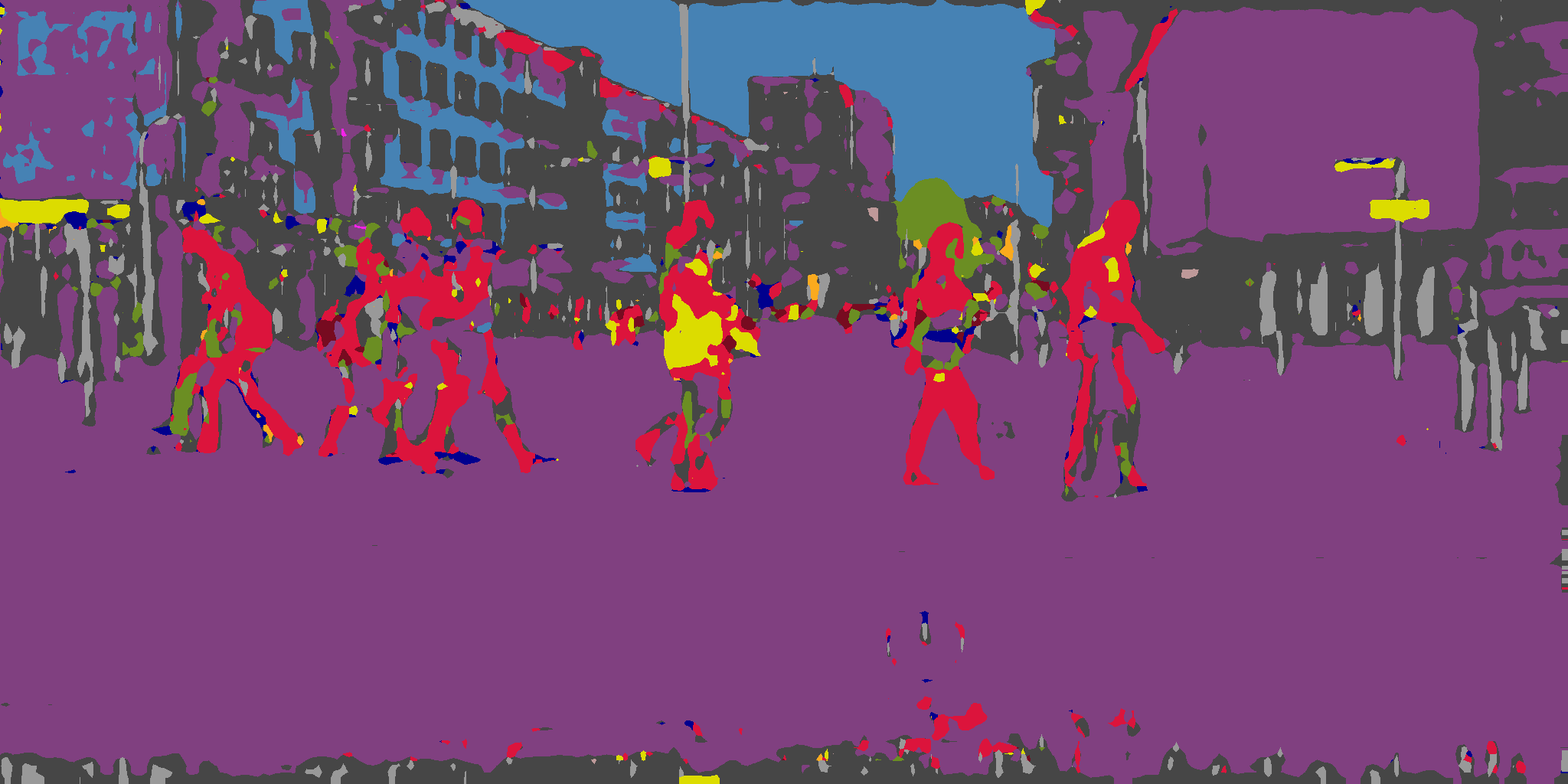}
    \\
    \includegraphics[width=0.32\textwidth]{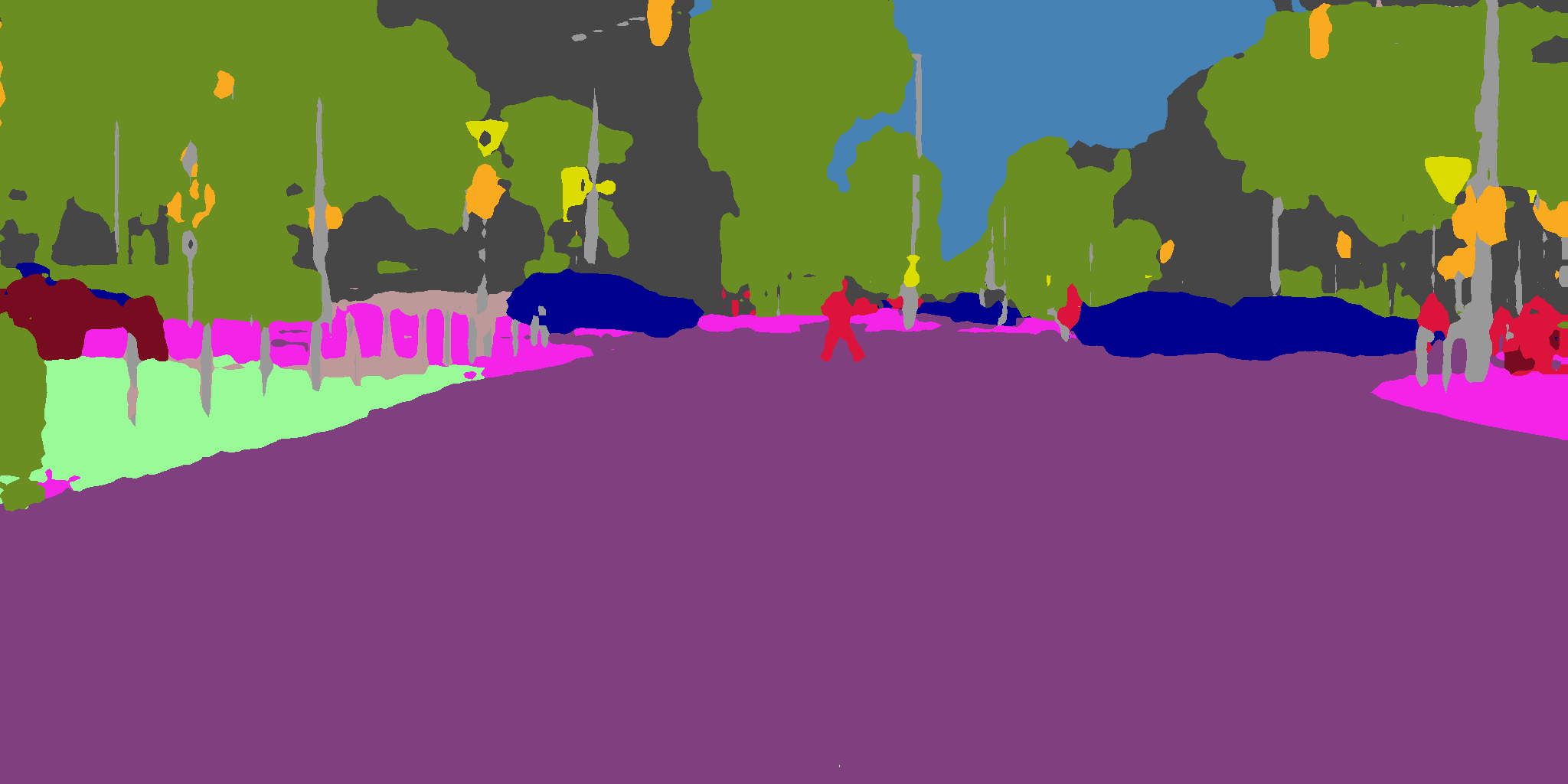}
    \includegraphics[width=0.32\textwidth]{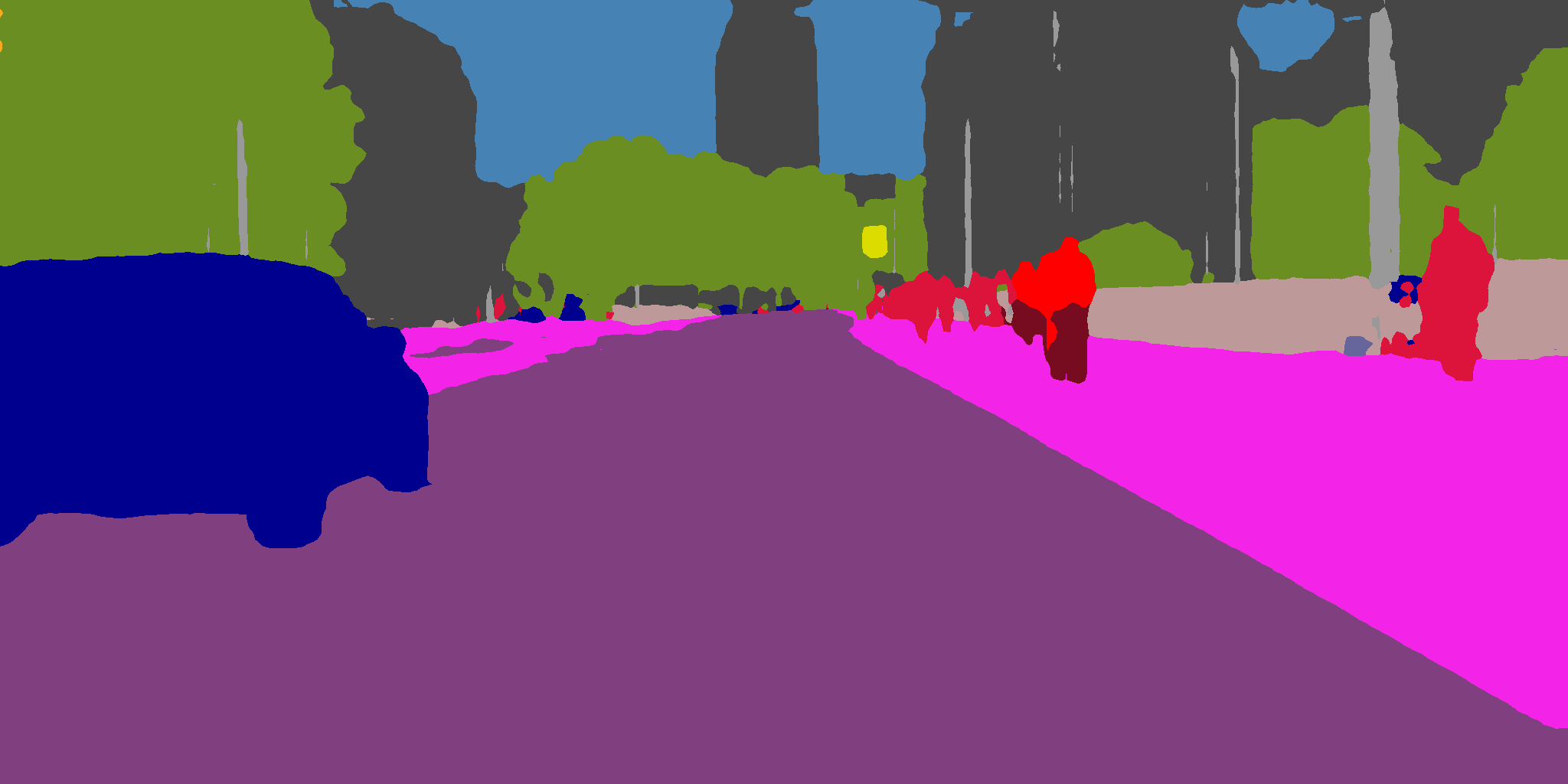}
    \includegraphics[width=0.32\textwidth]{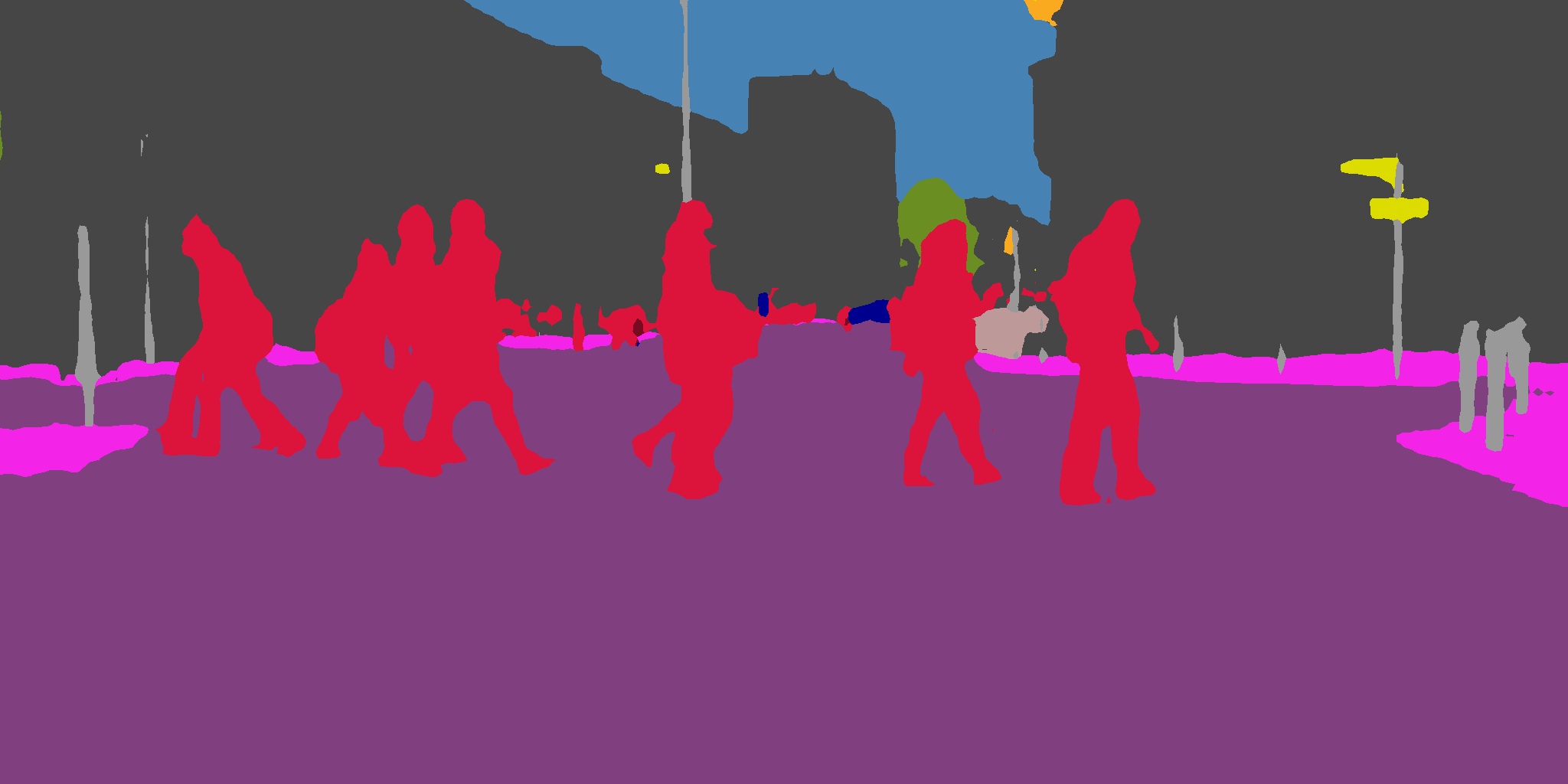}
\end{center}
\caption{Visual comparison on Cityscape validation set \cite{cityscaples2016}. First row: input RGB images; second row: ground truths; third row: context branch outputs; fourth row: detail branch outputs; and last row: ContextNet outputs using both context and detail branches. ContextNet obtained 65.9\% mIoU, while global context with one fourth resolution and detail branches trained independently obtained 63.3\% and 25.1\% mIoU respectively}
\label{fig:results-cityscapes-viz}
\end{figure}

\subsubsubsection{Context Branch Analysis}
We have zeroed the output of either the detail branch or the context branch to observe their individual contributions. The results are shown in Figure \ref{fig:results-cityscapes-viz}. In the first column, we see that the global context branch detects larger objects correctly (for example sky or trees) but fails around boundaries and thin regions. In contrast, the detail branch detects object boundaries correctly but fails at the centre region of objects. One exception is the centre region of trees, which is classified correctly, probably due to the discriminative nature of tree texture. Finally, we can see that ContextNet correctly combines both information to produce improved results in last row. Similarly, in the middle column we can see that segmentation of the pedestrians are refined by ContextNet over the global context branch output. In the last column, even though detail branch detects poles and traffic signs, ContextNet fails to effectively combine some of these with the global context branch. Overall we observe that the context branch and the detail branch learn complementary information.

\subsubsection{Cityscape Benchmark Results}
We evaluate ContextNet on the withheld test-set of Cityscapes \cite{cityscaples2016}. Table~\ref{tbl:results-cityscapes} shows the results in comparison to current state-of-the-art real-time segmentation networks (SegNet \cite{badrinarayanan2017}, ENet \cite{paszke2016}, ICNet \cite{zhao2017b} and ERFNet \cite{romera2018}), and offline methods (PSPNet \cite{zhao2017a} and DeepLab-v2 \cite{chen2016}). Table~\ref{tbl:results-cityscapes-time} compares the runtime at full, half and quarter resolution on images of Cityscapes. ContextNet achieves 64.2\% before, and 66.1\% mIoU after pruning (explained below), and runs at 18.3 fps in a single CPU thread of TensorFlow~\cite{tensorflow2015}. ENet~\cite{paszke2016} has a similar run-time but achieves only 58.3\% accuracy. ICNet \cite{zhao2017b} and ERFNet \cite{romera2018} achieve 69.5\% and 68.0\% mIoU, respectively, but are considerably slower than ContextNet.\footnote{Although our implementation of ICNet \cite{zhao2017b} achieves similar accuracy, we do not achieve the timing mentioned by the authors. This might be caused by differences in software environment and the employed testing protocols.}
\begin{table}[t]
\begin{center}
\scalebox{0.8}{%
    \begin{tabular}{|l|c|c|c|} \hline
    ~ & Class mIoU (in \%) & Category mIoU (in \%) & Parameters (in millions) \\ \hline
    DeepLab-v2 \cite{chen2016}* & 70.4 & 86.4 & 44.\_\_ \\ \hline
    PSPNet \cite{zhao2017a}* & \textbf{78.4} & \textbf{90.6} & 65.7\_ \\ \hline \hline
    SegNet \cite{badrinarayanan2017} & 56.1 & 79.8 & 29.46 \\ \hline
    ENet \cite{paszke2016} & 58.3 & 80.4 & 00.37 \\ \hline
    ICNet \cite{zhao2017b}* & 69.5 & - & 06.68 \\ \hline
    ERFNet \cite{romera2018} & 68.0 & 86.5 & 02.1\_ \\ \hline \hline
    ContextNet (Ours) & 66.1 & 82.7 & 00.85 \\ \hline
    \end{tabular}}
\end{center}
      \caption{Cityscape benchmark results for the proposed ContextNet and similar networks. DeepLab-v2~\cite{chen2016} and PSPNet~\cite{zhao2017a} are considered offline approaches. Runtime of other methods is shown in Table~\ref{tbl:results-cityscapes-time}. (Methods with `*' are pre-trained on ImageNet \cite{imagenet2015}.)}
  \label{tbl:results-cityscapes}
\end{table}
\begin{table}[t]
\begin{center}
\scalebox{0.8}{%
    \begin{tabular}{|l|c|c|c|} \hline
    ~ & $1024\times2048$  & $512\times1024$ & $256\times512$ \\ \hline
    SegNet \cite{badrinarayanan2017}* & 1.6 & - & - \\ \hline
    ENet \cite{paszke2016}* & 20.4 & 76.9 & 142.9 \\ \hline
    ICNet \cite{zhao2017b} & 14.2 & 46.3 & 83.2 \\ \hline
    ERFNet \cite{romera2018}* & 11.2 & 41.7 & 125.0\\ \hline
    ContextNet (Ours) & 18.3 & 65.5 & 139.2 \\ \hline
    ContextNet (Ours)$\dagger$ & 41.9 & 136.2 & 299.5 \\ \hline
    \end{tabular}}
\end{center}
      \caption{Runtime on Nvidia Titan X (Maxwell, 3,072 CUDA cores) with TensorFlow~\cite{tensorflow2015}, including sequential CPU/GPU memory transfer and kernel execution. (Results with `*' are taken from existing literature -- it is not known if memory transfer is considered. Our measure with `$\dagger$' denotes kernel execution time alone.)}
  \label{tbl:results-cityscapes-time}
\end{table}

We emphasize that our runtime evaluation includes the complete CPU and GPU pipeline including memory transfers. If parallel memory transfer and kernel execution are employed, our run time improves to 41.9 fps. Finally, in Table~\ref{tbl:results-cityscapes-time} we observe that ContextNet scales well for smaller input resolution sizes, and therefore can be tuned for the task at hand and the available resources. The results of ContextNet are displayed in Figure \ref{fig:results-cityscapes-viz} for qualitative analysis. ContextNet is able to segment even small objects at far distances adequately.

\subsubsubsection{Network Pruning} Network pruning is usually employed to reduce network parameters \cite{li2017,zhao2017b}. Following the ``lottery ticket'' intuition~\cite{frankle2018lottery} we start with wider networks (larger number of feature maps) and use pruning to obtain ``skinnier'' networks in order to increase the accuracy of our model. 
First, we train our network with twice the (target) number of feature maps to obtain improved results. We then decrease parameters progressively by pruning to 1.5, 1.25 and 1 times the original size. Following \cite{li2017}, we pruned filters with lowest $\ell_1$ sum. Via pruning we improve the mIoU from 64.2\% to 66.1\% on the Cityscape test set.

\section{Conclusions and Future Work}
\label{sec:conclusions}
In this work we proposed a real-time semantic segmentation model called ContextNet, which combines a deep network branch at low resolution but with large receptive field with a shallow and therefore efficient full-resolution branch to enhance the segmentation details.
ContextNet further extensively leverages depth-wise convolutions and bottleneck residual blocks for maximum memory and run-time efficiency.

Our ablation study shows that ContextNet effectively combines global and local context to achieve competitive result and outperforms other state-of-the-art real-time methods. We also empirically show that model pruning (in order to reach given targets of network size and real-time performance) leads to improved segmentation accuracy. Demonstrating that the ContextNet architecture is beneficial for other tasks relevant for autonomous systems such as single-image depth prediction is part of future work.
\bibliographystyle{splncs}
\bibliography{refs}
\end{document}